\documentclass[11pt]{article}

\usepackage[final]{acl}

\usepackage{times}
\usepackage{latexsym}

\usepackage[T1]{fontenc}

\usepackage[utf8]{inputenc}
\usepackage{tipa}
\usepackage{newunicodechar}
\DeclareUnicodeCharacter{0254}{\textopeno}
\DeclareUnicodeCharacter{0272}{\textltailn}
\DeclareUnicodeCharacter{0257}{\texthtd}
\DeclareUnicodeCharacter{0253}{\texthtb}
\DeclareUnicodeCharacter{03B1}{\alpha}
\DeclareUnicodeCharacter{025B}{\textepsilon}
\DeclareUnicodeCharacter{014B}{\ng}
\DeclareUnicodeCharacter{1ECD}{\d{o}}
\DeclareUnicodeCharacter{1EE5}{\d{u}}
\DeclareUnicodeCharacter{1ECB}{\d{i}}
\DeclareUnicodeCharacter{1EB9}{\d{e}}
\usepackage{microtype}

\usepackage{inconsolata}

\usepackage{graphicx}

\usepackage{amsmath}
\usepackage{amssymb}
\usepackage{algorithm}
\usepackage{algorithmic}
\usepackage{booktabs}
\usepackage{xcolor}
\usepackage{tcolorbox}
\usepackage{multirow}
\usepackage{graphicx}
\usepackage{subcaption}
\title{NSL-MT: Linguistically Informed Negative Samples for Efficient Machine Translation in African Low-Resource Languages}

\author{
 \textbf{Mamadou K. KEITA\textsuperscript{1}},
  \textbf{Christopher Homan\textsuperscript{1}},
 \textbf{Huy Le \textsuperscript{1}}
\\
\textsuperscript{1}Rochester Institute of Technology
}

\begin{document}
\maketitle
\begin{abstract}
We introduce negative space learning machine translation (NSL-MT), a training method for  underresourced languages, that augments limited parallel data with synthetically generated violations of the target language's grammar and explicitly penalizes the model when it assigns high probability to these linguistically invalid outputs.  NSL-MT delivers improvements across all baselines we tested, including 3-12\% BLEU gains for well-performing models and 56-89\% gains for models lacking decent initial support. Furthermore, NSL-MT provides a 5x data efficiency multiplier: training with 1,000 examples matches or exceeds normal training with 5,000 examples. NSL-MT thus provides a data-efficient alternative training method for settings where parallel data is limited.
\end{abstract}

\section{Introduction}
\label{sec:intro}

\begin{figure}[h]
    \centering    \includegraphics[width=8cm]{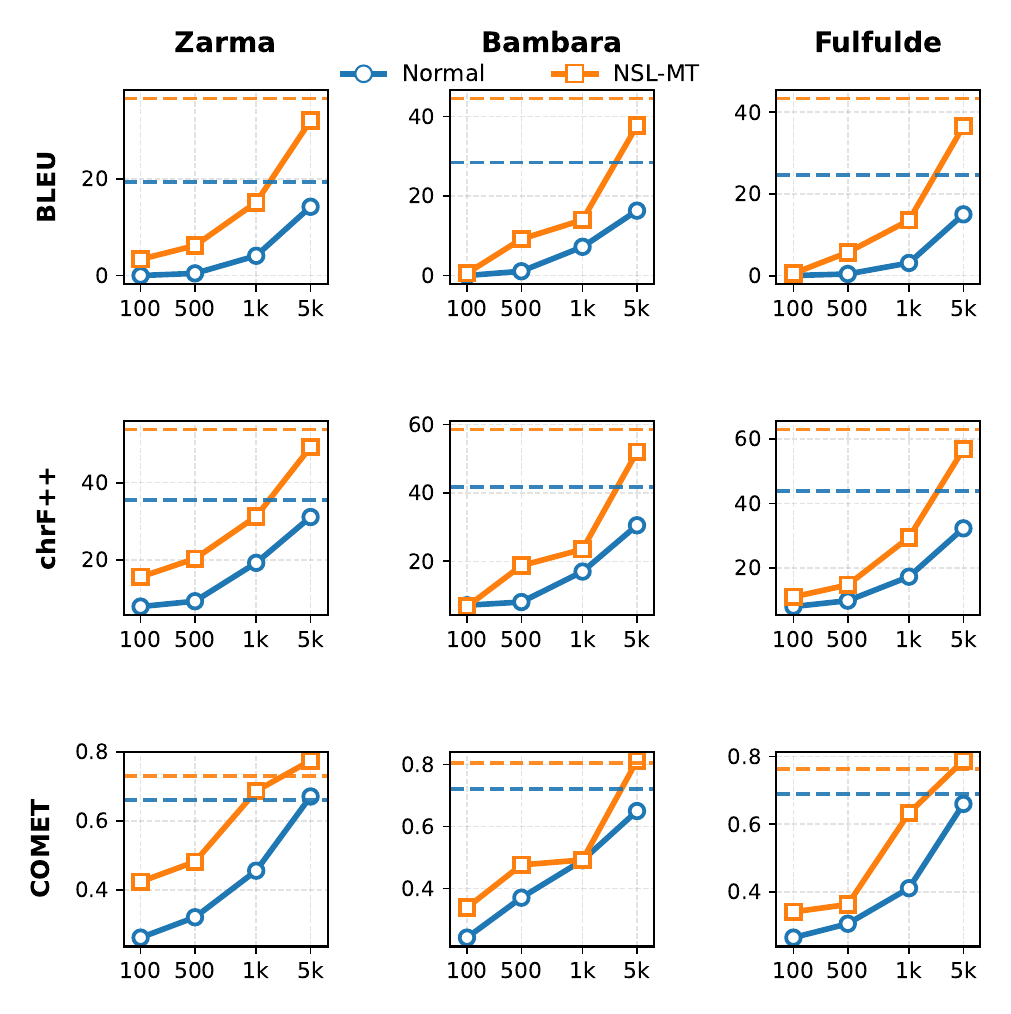}
    \caption{Data efficiency comparison between Normal training and NSL-MT across varying training set sizes. NSL-MT achieves high performance at all data sizes, with the largest relative gains occurring at the smallest data sizes.}
    \label{fig:data-efficiency}
\end{figure}

For high-resource languages, where millions of parallel sentences are available for training, neural machine translation (MT) has  in recent years seen remarkable advances \citep{Zhang2020}. Unfortunately, the majority of the world's 7K+ languages lack such abundant training resources, with 15K sentence pairs being the practical limit.  For these \emph{low-resource} languages, collecting parallel data is expensive \citep{magueresse2020lowresourcelanguagesreviewpast}.
Yet linguistic expertise often exists in the form of native speakers who can articulate grammar rules. This condition reflects the reality for hundreds of African, indigenous, and minority languages worldwide. Moreover, many of these languages exist in sociolinguistic contexts where high-resource `colonial' languages dominate official communication and education~\citep{10.18356/27081990-151:/content/papers/10.18356/27081990-151}, which leaves large populations, often with limited literacy, unable to access important information in their native languages~\footnote{This motivates our focus on translating \emph{from} high-resource languages \emph{to} low-resource languages, rather than the reverse, as such translation enables broader information access. 
}.

In this research we ask: \emph{\textbf{can we leverage explicit linguistic knowledge to compensate for scarce parallel data?}}


In conventional neural MT training, models can, with enough data, learn boundaries between acceptable and unacceptable outputs by observing distributional patterns. In low-resource settings, this implicit learning fails. Models encounter too few examples to reliably distinguish grammatical patterns from noise. This results in characteristic errors, such as source language word order imposed on, morphology applied to, or function words inserted into the target language.

To overcome these challenges, we propose \emph{negative space learning} machine translation (NSL-MT), an approach
that explicitly teaches models what \textit{\textbf{not}} to generate. 
NSL-MT composes two innovations:

First, it generates linguistically guided hard negative examples. Contrastive learning methods \citep{chen2020simple,gao2021simcse} typically sample negatives from the training distribution or apply generic augmentation. These negatives are often far from decision boundaries, and provide weak learning signal. By contrast, our  examples are generated to fall closer to the decision boundary and exhibit the characteristic errors found in low-resource MT.

Second, it modifies the training objective rather than the (positive) training data distribution. Data augmentation methods like back-translation or paraphrasing create additional positive examples while maintaining standard maximum likelihood objectives \citep{Qumar2025,mallinson2017paraphrasing,sennrich2016improving}.


We contribute:
\begin{itemize}
\item NSL-MT, a training approach that encodes linguistic constraints as severity-weighted penalties in the loss function, teaching models to avoid linguistically invalid outputs. \textbf{All codes will be open-sourced upon acceptance.}

\item Experiments showing, among other things, that NSL-MT:

\begin{itemize}
    \item provides consistent improvements across four model architectures with  gains as high as 89\% in BLEU score for models that lack initial support for the target languages, and substantial if more modest gains for other models.
 \item offers a 5x data efficiency multiplier, i.e., training with NSL-MT on 1K examples matches or exceeds normal training on 5K examples.
\end{itemize}
\end{itemize}

\section{Related Work}
\label{sec:related}
Early approaches to low-resource machine translation focused on transfer learning from high-resource languages \citep{zoph2016transfer} or multilingual training that shares representations across language families \citep{aharoni2019massively}. These methods improve over monolingual baselines but struggle when source and target languages differ typologically~\citep{muller2022languagesknowinfluencelearn}. Our work addresses this limitation by explicitly encoding cross-lingual structural differences as negative constraints.

Recent multilingual pre-trained models demonstrate good cross-linguage transfer. The mT5 model \citep{xue2021mt5} pre-trains on 101 languages using a masked span prediction objective, while NLLB \citep{nllb2022} trains on 200 languages using a combination of parallel data and back-translation. AfriMT5 \citep{adelani2022few} specializes multilingual pre-training for African languages. These models provide strong starting points, but still require fine-tuning to achieve good performance on low-resource languages. We demonstrate that NSL-MT improves fine-tuning efficiency across different model architectures.

Data augmentation methods create synthetic training examples to increase effective dataset size. Back-translation generates source sentences from target monolingual data, while paraphrasing creates alternative translations of existing parallel sentences \citep{Qumar2025,mallinson2017paraphrasing,sennrich2016improving}. These approaches alter the training distribution and maintain standard maximum likelihood objectives. NSL-MT differs fundamentally by modifying the training objective itself to include negative evidence. Furthermore, back-translation requires \textit{strong reverse-direction} models that do not exist for most low-resource languages. NSL-MT requires only linguistic knowledge, which native speakers can provide.

Contrastive learning has proven effective for representation learning and natural language processing \citep{gao2021simcse,chen2020simple}. These methods learn by distinguishing positive from negative examples in the embedding space. 
They are particularly effective at learning human preferences, but doing so requires feedback from humans~\citep{hejna2024contrastivepreferencelearninglearning}, a relatively expensive resource.
NSL-MT shares the core principle of learning from negative evidence but operates at the sentential rather than the representation level. Where contrastive methods generate negatives through random sampling or data augmentation, NSL-MT relies on targeted linguistic violations, creating hard negatives that exclusively address known failure modes. This distinction proves useful for low-resource settings, where random negatives prove insufficient or detrimental to the learning signal.

Reinforcement learning from human feedback (RLHF) is widely used for aligning language models with human preferences \citep{ouyang2022training}. RLHF methods, like the direct preference optimization (DPO) or  $\Phi$PO~\citep{rafailov2024directpreferenceoptimizationlanguage, azar2023generaltheoreticalparadigmunderstand}, train reward models on human judgments of output quality, then uses these rewards to fine-tune generation models. This approach shares NSL-MT's goal of teaching models what not to generate. However, RLHF requires collecting human judgments at scale, which is relatively expensive, especially for low-resource languages. NSL-MT provides a practical alternative by encoding linguistic constraints that would be \emph{expensive} to learn from human feedback. Where RLHF learns from implicit human preferences, NSL-MT encodes explicit linguistic rules.

Work on cross-lingual transfer has investigated how linguistic typology affects transfer success. Studies show that syntactic similarity predicts transfer effectiveness better than relatedness or geographical proximity \citep{blaschke2025analyzingeffectlinguisticsimilarity,lin2019choosing}. This finding aligns our constraint type ablation study, which reveals that different violation categories contribute differently across languages based on their typological properties. Our results confirm that effective transfer requires explicit attention to structural differences between source and target languages.

Instruction tuning research has shown that task descriptions and demonstrations improve models (language models) performance on downstream tasks \citep{wei2022finetuned}. This work demonstrates the value of explicit task specification beyond implicit pattern learning. NSL-MT applies similar principles to translation quality by explicitly specifying what constitutes incorrect outputs. While instruction tuning tells models what to do, NSL-MT tells models what \textit{not} to also do.

\section{NSL-MT}
\label{sec:nsl}

NSL-MT is a training approach, and an implementation of the Principled Learning (PrL) paradigm \cite{keita2025r2truleencodedlossfunctions}, that explicitly teaches translation models what \textit{\textbf{not}} to generate. NSL-MT augments parallel data with synthetically generated negative examples that violate specific linguistic constraints. The model learns to assign lower probability to these violations, thereby improving its understanding of correct target language structure.

\paragraph{Core principle:} Most of the neural MT trainings optimize the model to maximize the likelihood of correct translations. 
 However, maximum likelihood estimation alone provides no explicit indication about what the model should avoid. NSL-MT addresses this gap by introducing a contrastive objective that penalizes the model for assigning high probability to linguistically invalid outputs.

\subsection{Violation Generation}

For each parallel sentence pair $(x, y)$ in the training set, we generate a set of negative examples $\mathcal{V}(y) = \{v_1, v_2, \ldots, v_k\}$ where each $v_i$ is a corrupted version of $y$ that violates specific grammatical or structural rules of the target language. We define three categories of violations:

\textbf{Morphological violations} corrupt the internal structure of words. For languages without grammatical gender, we add a high resource language (\textbf{French in our case}\footnote{French is chosen because the languages covered in our study co-exist with it. The countries that speak these languages have French as part or only official language.})style gender markers. For languages with noun class systems, we substitute incorrect class affixes. For plural formation, we replace language-specific markers with French plural \textit{-s}. These violations target agreement patterns that low-resource models often fail to learn from positive examples alone.

\textbf{Syntactic violations} modify order and structural relationships. We apply word order transformations that impose French SVO patterns on SOV or VSO languages. We move adjectives from their correct post-nominal position to the pre-nominal position common in French. We replace postpositions with French prepositions. These violations prevent the model from defaulting to source language syntax.

\textbf{Lexical violations} introduce inappropriate vocabulary choices. We insert French articles where the target language uses suffixes. We add French auxiliary verbs where the target language uses different marking systems. We apply French negation patterns instead of target language negation approaches. These violations address interference from high-resource source language patterns.

Each violation type $t$ carries an associated severity weight $s_t \in [0,1]$ that reflects its impact on comprehension. We assign higher severity to violations that fundamentally break grammatical agreement ($s_t = 1.0$) and lower severity to violations that affect style but preserve basic meaning ($s_t = 0.6$).

\subsection{Training Objective}
NSL-MT combines positive and negative learning signals in a unified objective. For a training batch containing both correct translations and generated violations, we compute
\begin{equation*}
\mathcal{L}_{\text{NSL-MT}} = \mathcal{L}_{\text{pos}} + \alpha \mathcal{L}_{\text{neg}}
\end{equation*}
where 
$\alpha$ is a weighting hyperparameter.

We define the positive loss as usual:
\begin{equation*}
\mathcal{L}_{\text{pos}} = -\sum_{(x,y) \in \mathcal{D}_{\text{pos}}} \log P(y|x;\theta)
\end{equation*}
where $\theta$ represents model parameters.

For negative examples, we want the model to assign \textbf{low} probability to linguistically invalid outputs. This is done by \textbf{minimizing the (severity-weighted) log-probability of violations}:
\begin{equation*}
\mathcal{L}_{\text{neg}} = \sum_{(x,v) \in \mathcal{D}_{\text{neg}}} s(v) \cdot \log P(v|x;\theta)
\end{equation*}
where $s(v)$ is the severity weight of violation $v$. Higher severity means a stronger penalty when the model likes a bad output~\footnote{In the official implementation, we use $(1 + \alpha \cdot s(v)) \cdot \text{CE}(v)$ instead of $s(v) \cdot \log P(v \mid x)$ directly, as it keeps the total loss positive and numerically stable in PyTorch. This is similar to the unlikelihood training of \citet{welleck2019neuraltextgenerationunlikelihood}.}.

\subsection{Implementation}

\begin{algorithm}[t]
\tiny
\caption{NSL-MT Training}
\label{alg:nsl}
\begin{algorithmic}[1]
\REQUIRE Training data $\mathcal{D} = \{(x_i, y_i)\}_{i=1}^N$, violation generators $\mathcal{G}$, model $M_\theta$
\REQUIRE Hyperparameters: $\alpha$ (negative weight), $\beta$ (learning rate), $K$ (epochs)
\FOR{epoch = 1 to $K$}
    \STATE Initialize batch $\mathcal{B}_{\text{pos}} \leftarrow \emptyset$, $\mathcal{B}_{\text{neg}} \leftarrow \emptyset$
    \FOR{each $(x, y) \in \mathcal{D}$}
        \STATE Add $(x, y)$ to $\mathcal{B}_{\text{pos}}$
        \STATE $k \sim \text{Uniform}(3, 5)$ \COMMENT{Sample number of violations per correct example}
        \FOR{$j = 1$ to $k$}
            \STATE $t \sim \text{Uniform}(\mathcal{G})$ \COMMENT{Sample violation type}
            \STATE $v_j, s_j \leftarrow \mathcal{G}_t(y)$ \COMMENT{Generate violation + severity}
            \STATE Add $(x, v_j, s_j)$ to $\mathcal{B}_{\text{neg}}$
        \ENDFOR
    \ENDFOR
    \STATE $\mathcal{L}_{\text{pos}} \leftarrow -\frac{1}{|\mathcal{B}_{\text{pos}}|} \sum_{(x,y) \in \mathcal{B}_{\text{pos}}} \log P(y|x;\theta)$ \COMMENT{standard CE on correct translations}
    \STATE $\mathcal{L}_{\text{neg}} \leftarrow \frac{1}{|\mathcal{B}_{\text{neg}}|} \sum_{(x,v,s) \in \mathcal{B}_{\text{neg}}} s \cdot \log P(v|x;\theta)$ \COMMENT{push down probability of violations}.
    \STATE $\mathcal{L} \leftarrow \mathcal{L}_{\text{pos}} + \alpha \mathcal{L}_{\text{neg}}$
    \STATE $\theta \leftarrow \theta - \beta \nabla_\theta \mathcal{L}$ \COMMENT{Update parameters}
\ENDFOR
\end{algorithmic}
\end{algorithm}

Algorithm~\ref{alg:nsl} presents the NSL-MT training procedure. For each training example, we generate violations to prevent the model from memorizing specific corrupted sequences. We sample the number of violations uniformly between 3 and 5 per positive example, creating a 3:1 to 5:1 ratio of negative to positive examples.

We implement violation generators as rule-based systems that encode linguistic knowledge about common error patterns. Each generator takes a correct target sentence $y$ and produces a corrupted version $v$ along with its severity score $s$. The generators operate the same way for a given input and violation type, but we introduce randomness by sampling which violations to apply and in which order.

During training, we shuffle positive and negative examples within each batch to prevent the model from learning position-based patterns. We apply standard techniques such as gradient clipping and learning rate warmup to ensure stable optimization.

\section{Experiments and Results}
\label{sec:experiments}

We investigate the following questions.

\textbf{RQ1.} Do explicit negative rules encoded in the loss function improve low-resource translation quality compared to standard maximum likelihood training?

\textbf{RQ2.} How does NSL-MT's effectiveness vary with training data size, and at what threshold does NSL-MT provide maximum relative benefit?

\textbf{RQ3.} Which types of linguistic constraints contribute most to translation quality?

We design our experiments to reflect realistic low-resource translation scenarios~\footnote{By low resource scenario, we are not referring to limited number of speakers, but the availability of resources. Therefore, there is no need to test on high-resource African languages. The selected languages are truly low-resource.}. We consider a setting where annotated parallel data is limited (at most 15,000 sentence pairs) and the cost of creating additional parallel corpora is expensive. However, bilingual speakers who understand both the source and target languages can gather grammatical rules and common error patterns in a matter of hours.

To validate NSL-MT under these conditions, we select three languages spoken in West Africa: Zarma (Nilo-Saharan family), Bambara (Mande family), and Fulfulde (Atlantic-Congo family). \textbf{We further tested on English to African languages} (see Section \ref{sec:eng_afri}).


\subsection{Experimental Setup}

\paragraph{Datasets} We use multi-domain instruction dataset, InstructLR \citep{keita-etal-2026-instructlr}. The dataset has 3 benchmarks of 50,000 examples for each of the our three languages and every instruction has its french translation. For each language, we extract 15,000 instruction sentence pairs for training, 500 pairs for validation, and 1,000 pairs for testing. We ensure no overlap between splits and all the models are trained on the same selected sets.

\paragraph{Models} We evaluate NSL-MT on four multilingual translation models that cover different architectures:

\textbf{NLLB-200-distilled-600M} \citep{nllb2022}: A 600M parameter model trained on 200 languages We fine-tune the distilled version on our target languages.

\textbf{AfriMT5-base} \citep{adelani2022few}: A 300M parameter encoder-decoder model pre-trained on 17 African languages.

\textbf{mT5-base} \citep{xue2021mt5}: A 580M parameter multilingual variant of T5 pre-trained on 101 languages using masked language modeling.

\textbf{mT5-small}: A 300M parameter version of mT5. We include this model to test NSL-MT effectiveness on smaller architectures.

The training configurations are detailed in section~\ref{sec:exp_configs}.

\subsection{Results}
\label{sec:results}
Table~\ref{tab:main-results} presents the performance of NSL-MT compared to standard training across all models and languages. NSL-MT outperforms normal training across all evaluation metrics and model architectures.

\begin{table*}[t]
\centering
\tiny
\resizebox{\textwidth}{!}{
\begin{tabular}{ll|ccc|ccc|ccc}
\multirow{2}{*}{\textbf{Model}} & \multirow{2}{*}{\textbf{Method}} & \multicolumn{3}{c|}{\textbf{Zarma}} & \multicolumn{3}{c|}{\textbf{Bambara}} & \multicolumn{3}{c}{\textbf{Fulfulde}} \\
& & BLEU & chrF++ & COMET & BLEU & chrF++ & COMET & BLEU & chrF++ & COMET \\
\midrule
\multirow{3}{*}{NLLB-200}
& Normal & $38.33_{\pm 1.2}$ & $56.63_{\pm 0.8}$ & 0.7976 & $53.27_{\pm 1.5}$ & $67.03_{\pm 0.9}$ & 0.8617 & $47.61_{\pm 1.3}$ & $67.56_{\pm 0.7}$ & 0.8211 \\
& NSL-MT & $\mathbf{42.82_{\pm 1.1}}$ & $\mathbf{60.67_{\pm 0.7}}$ & \textbf{0.8078} & $\mathbf{55.11_{\pm 1.4}}$ & $\mathbf{68.40_{\pm 0.8}}$ & \textbf{0.8637} & $\mathbf{49.31_{\pm 1.2}}$ & $\mathbf{70.04_{\pm 0.6}}$ & \textbf{0.8264} \\
& $\Delta$ (\%) & \textit{+11.7} & \textit{+7.1} & \textit{+1.3} & \textit{+3.5} & \textit{+2.0} & \textit{+0.2} & \textit{+3.6} & \textit{+3.7} & \textit{+0.6} \\
\midrule
\multirow{3}{*}{AfriMT5-Base}
& Normal & $19.36_{\pm 1.5}$ & $35.44_{\pm 1.2}$ & 0.6608 & $28.44_{\pm 1.8}$ & $41.68_{\pm 1.3}$ & 0.7202 & $24.58_{\pm 1.6}$ & $43.88_{\pm 1.1}$ & 0.6887 \\
& NSL-MT & $\mathbf{36.62_{\pm 1.3}}$ & $\mathbf{53.78_{\pm 0.9}}$ & \textbf{0.7311} & $\mathbf{44.51_{\pm 1.6}}$ & $\mathbf{58.51_{\pm 1.0}}$ & \textbf{0.8063} & $\mathbf{43.33_{\pm 1.4}}$ & $\mathbf{62.94_{\pm 0.8}}$ & \textbf{0.7629} \\
& $\Delta$ (\%) & \textit{+89.2} & \textit{+51.8} & \textit{+10.6} & \textit{+56.5} & \textit{+40.4} & \textit{+12.0} & \textit{+76.3} & \textit{+43.4} & \textit{+10.8} \\
\midrule
\multirow{3}{*}{mT5-Base}
& Normal & $7.67_{\pm 2.1}$ & $17.80_{\pm 1.8}$ & 0.4658 & $0.05_{\pm 0.3}$ & $3.83_{\pm 1.5}$ & 0.3302 & $5.02_{\pm 1.9}$ & $16.34_{\pm 1.6}$ & 0.3843 \\
& NSL-MT & $\mathbf{33.21_{\pm 1.4}}$ & $\mathbf{50.71_{\pm 1.0}}$ & \textbf{0.7221} & $\mathbf{41.35_{\pm 1.5}}$ & $\mathbf{55.11_{\pm 1.1}}$ & \textbf{0.7935} & $\mathbf{41.14_{\pm 1.5}}$ & $\mathbf{60.02_{\pm 0.9}}$ & \textbf{0.7535} \\
& $\Delta$ (\%) & \textit{+333.1} & \textit{+184.9} & \textit{+55.0} & \textit{+82600} & \textit{+1339.4} & \textit{+140.3} & \textit{+719.5} & \textit{+267.4} & \textit{+96.1} \\
\midrule
\multirow{3}{*}{mT5-Small}
& Normal & $0.06_{\pm 0.4}$ & $3.39_{\pm 1.6}$ & 0.2875 & $0.02_{\pm 0.2}$ & $3.42_{\pm 1.4}$ & 0.2652 & $0.03_{\pm 0.3}$ & $3.80_{\pm 1.5}$ & 0.3059 \\
& NSL-MT & $\mathbf{19.33_{\pm 1.7}}$ & $\mathbf{34.53_{\pm 1.3}}$ & \textbf{0.6577} & $\mathbf{23.42_{\pm 1.8}}$ & $\mathbf{35.79_{\pm 1.4}}$ & \textbf{0.6835} & $\mathbf{25.05_{\pm 1.7}}$ & $\mathbf{44.48_{\pm 1.2}}$ & \textbf{0.6831} \\
& $\Delta$ (\%) & \textit{+32116.7} & \textit{+918.9} & \textit{+128.8} & \textit{+116999} & \textit{+947.1} & \textit{+157.8} & \textit{+83400} & \textit{+1070.5} & \textit{+123.3} \\
\end{tabular}
}
\caption{Main results comparing baseline training and NSL-MT across four model architectures on three African languages. We train all models on 15,000 parallel sentences. $\Delta$ shows relative improvement ( $\frac{\text{NSL-MT} - \text{baseline}}{\text{baseline}} \times 100\%$).}
\label{tab:main-results}
\end{table*}

NSL-MT delivers improvements across all model architectures. For NLLB-200, which already performs well due to its large-scale pre-training, NSL-MT provides modest but gains from 3.5\% to 11.7\% BLEU improvement. For AfriMT5-base, which shows moderate baseline performance, NSL-MT yields improvements ranging from 56.5\% to 89.2\% BLEU improvement. For mT5-base and mT5-small, which struggle on these low-resource languages without NSL-MT, the improvements are higher.

The magnitude of improvement correlates inversely with baseline performance. Models that already translate reasonably well benefit less from NSL-MT, while models that produce poor translations without NSL-MT show gains. 

Across metrics, BLEU shows the largest relative improvements, followed by chrF++, and then COMET. BLEU measures exact n-gram matches and thus proves particularly sensitive to grammatical errors that NSL-MT targets. chrF++ operates at the character level and shows smaller but still improvements. COMET, which relies on learned semantic representations, shows consistent but more modest gains.

\paragraph{Human Evaluation} 

Two volunteer native speakers each for Zarma and Bambara blindly evaluated 50 random samples. For each sample, annotators selected their preferred translation, baseline or NSL-MT outputs, and rated confidence on a 1-5 scale (1=not sure, 5=very sure).

Table~\ref{tab:human-eval} shows that, for each language, both annotators preferred NSL-MT outputs on all samples (Cohen's $\kappa = 1.000$, $p < 0.001$). Mean confidence ratings exceeded 4.0/5.0 for both languages, with inter-annotator agreement of 79.5\% for exact matches and 93.2\% within $\pm$1 point. The perfect coherence score across all the languages is justifiable by the fact that with limited samples, the baselines produced very "incorrect" outputs, sometimes even unreadable; whereas NSL-MT produced outputs, although not totally correct, acceptable compare to the baseline models.

\begin{table}[h]
\centering
\scriptsize
\begin{tabular}{lccc}
\textbf{Language} & \textbf{Pref. NSL-MT} & \textbf{Avg. Confidence} & \textbf{Cohen's $\kappa$} \\
\midrule
Zarma & 100\% (50/50) & $4.09 \pm 0.38$ & $1.000$ *** \\
Bambara & 100\% (50/50) & $4.00 \pm 0.29$ & $1.000$*** \\
\end{tabular}
\caption{Human evaluation showing preference for NSL-MT over NLLB baseline across 50 samples per language, with high confidence ratings. Inter-annotator exact agreement: 79.5\%, within $\pm$1: 93.2\%. *** $p < 0.001$}
\label{tab:human-eval}
\end{table}

\subsection{Ablation Study}

We conduct an ablation study to determine which categories of linguistic constraints contribute most to NSL-MT performance. We train AfriMT5-base models using only morphological violations, only syntactic violations, or only lexical violations, and compare these to the full NSL-MT approach that combines all three categories.

Table~\ref{tab:constraint-ablation} presents the results. Each constraint type individually outperforms the baseline, confirming that all three categories provide useful learning signal. However, the relative contribution varies by language.

\begin{table*}[t]
\centering
\tiny
\begin{tabular}{l|ccc|ccc|ccc}
\multirow{2}{*}{\textbf{Constraint Type}} & \multicolumn{3}{c|}{\textbf{Zarma}} & \multicolumn{3}{c|}{\textbf{Bambara}} & \multicolumn{3}{c}{\textbf{Fulfulde}} \\
& BLEU & chrF++ & COMET & BLEU & chrF++ & COMET & BLEU & chrF++ & COMET \\
\midrule
Baseline (Normal) & $19.58_{\pm 1.5}$ & $35.55_{\pm 1.2}$ & 0.7158 & $27.56_{\pm 1.8}$ & $41.16_{\pm 1.3}$ & 0.7571 & $24.59_{\pm 1.6}$ & $43.73_{\pm 1.1}$ & 0.7302 \\
\midrule
Morphological Only & $31.21_{\pm 1.4}$ & $48.40_{\pm 1.0}$ & 0.7722 & $35.30_{\pm 1.7}$ & $49.84_{\pm 1.1}$ & 0.8042 & $\mathbf{38.46_{\pm 1.5}}$ & $\mathbf{58.14_{\pm 0.9}}$ & 0.7958 \\
Syntactic Only & $26.60_{\pm 1.5}$ & $43.53_{\pm 1.1}$ & 0.7532 & $\mathbf{39.27_{\pm 1.6}}$ & $\mathbf{53.34_{\pm 1.0}}$ & $\mathbf{0.8229}$ & $25.28_{\pm 1.6}$ & $44.43_{\pm 1.1}$ & 0.7331 \\
Lexical Only & $\mathbf{32.02_{\pm 1.3}}$ & $\mathbf{50.26_{\pm 0.9}}$ & $\mathbf{0.7714}$ & $31.22_{\pm 1.7}$ & $46.13_{\pm 1.2}$ & 0.7867 & $36.09_{\pm 1.5}$ & $57.54_{\pm 0.9}$ & $\mathbf{0.7914}$ \\
\midrule
Full NSL-MT & $36.12_{\pm 1.3}$ & $53.27_{\pm 0.9}$ & 0.7857 & $44.34_{\pm 1.6}$ & $58.41_{\pm 1.0}$ & 0.8406 & $43.41_{\pm 1.4}$ & $62.69_{\pm 0.8}$ & 0.8069 \\
\end{tabular}
\caption{Ablation study showing the contribution of different constraint types. We train all models on 15,000 parallel sentences using AfriMT5-base. Each row represents training with only the specified constraint type, except Full NSL-MT which uses all constraints.}
\label{tab:constraint-ablation}
\end{table*}

For Zarma, lexical violations provide the largest individual contribution (+12.4 BLEU), followed by morphological violations (+11.6 BLEU) and syntactic violations (+7.0 BLEU). This pattern reflects Zarma's reliance on particles and auxiliaries, confirmed by our rule set.

For Bambara, syntactic violations dominate (+11.7 BLEU), outperforming morphological (+7.7 BLEU) and lexical (+3.7 BLEU) violations.

For Fulfulde, morphological violations contribute the most (+13.9 BLEU), lexical violations also help (+11.5 BLEU), while syntactic violations provide minimal benefit (+0.7 BLEU).

The full NSL-MT approach that combines all constraint types outperforms any single category, with gains ranging from 4.1 to 7.3 BLEU over the best individual constraint type. This additive effect indicates that different violation categories capture complementary aspects of linguistic features.

\subsection{Data Efficiency Analysis}

We investigate how NSL-MT performance scales with training data size. We train AfriMT5-base models using 100, 500, 1,000, and 5,000 parallel sentences with both normal training and NSL-MT. Figure~\ref{fig:data-efficiency} plots the learning curves for all three languages across all three metrics.

NSL-MT outperforms normal training at every data size across all languages and metrics. The advantage of NSL-MT increases as data becomes limited. At 100 examples, normal training produces near-zero BLEU scores (0.01-0.04), while NSL-MT achieves 0.55-3.38 BLEU, representing gains of 0.5-3.3 points. At 500 examples, NSL-MT provides gains of 5.7-8.1 BLEU points. At 1,000 examples, NSL-MT achieves 13.55-15.15 BLEU compared to 3.12-7.23 for normal training, representing improvements of 8.3-11.0 points. At 5,000 examples, NSL-MT maintains substantial advantages with gains of 17.8-21.4 BLEU points.

Table~\ref{tab:data-efficiency} quantifies the data efficiency of NSL-MT by comparing performance at different data sizes. NSL-MT with 1,000 examples matches or exceeds normal training with 5,000 examples for Zarma across all metrics. For Bambara and Fulfulde, NSL-MT with 1,000 examples achieves 76-90\% of normal training performance with 5,000 examples. This finding demonstrates that NSL-MT provides a \textbf{5x data efficiency multiplier} in practical terms.

\begin{table*}[h]
\centering
\tiny
\resizebox{\textwidth}{!}{
\begin{tabular}{ll|ccc|ccc|ccc}
& & \multicolumn{3}{c|}
{\textbf{Zarma}} & \multicolumn{3}{c|}{\textbf{Bambara}} & \multicolumn{3}{c}{\textbf{Fulfulde}} \\
\textbf{Data Size} & \textbf{Method} & BLEU & chrF++ & COMET & BLEU & chrF++ & COMET & BLEU & chrF++ & COMET \\
\midrule
\multirow{2}{*}{100} 
& Normal & 0.02$_{\pm 0.4}$ & 7.92 & 0.262 & 0.01$_{\pm 0.2}$ & 7.15 & 0.241 & 0.04$_{\pm 0.3}$ & 8.03 & 0.265 \\
& NSL-MT & \textbf{3.38}$_{\pm 2.0}$ & \textbf{15.61} & \textbf{0.424} & \textbf{0.58}$_{\pm 1.1}$ & \textbf{6.76} & \textbf{0.338} & \textbf{0.55}$_{\pm 1.2}$ & \textbf{11.04} & \textbf{0.341} \\
\midrule
\multirow{2}{*}{500}
& Normal & 0.47$_{\pm 1.5}$ & 9.31 & 0.321 & 1.09$_{\pm 1.3}$ & 8.09 & 0.370 & 0.42$_{\pm 1.4}$ & 9.86 & 0.306 \\
& NSL-MT & \textbf{6.21}$_{\pm 1.8}$ & \textbf{20.33} & \textbf{0.482} & \textbf{9.18}$_{\pm 1.6}$ & \textbf{18.73} & \textbf{0.476} & \textbf{5.71}$_{\pm 1.7}$ & \textbf{14.75} & \textbf{0.363} \\
\midrule
\multirow{2}{*}{1,000}
& Normal & 4.11$_{\pm 1.9}$ & 19.25 & 0.456 & 7.23$_{\pm 1.9}$ & 16.99 & 0.491 & 3.12$_{\pm 1.8}$ & 17.30 & 0.411 \\
& NSL-MT & \textbf{15.15}$_{\pm 1.7}$ & \textbf{31.27} & \textbf{0.687} & \textbf{14.00}$_{\pm 1.8}$ & \textbf{23.50} & \textbf{0.492} & \textbf{13.55}$_{\pm 1.7}$ & \textbf{29.47} & \textbf{0.633} \\
\midrule
\multirow{2}{*}{5,000}
& Normal & 14.24$_{\pm 1.7}$ & 31.12 & 0.671 & 16.33$_{\pm 1.8}$ & 30.53 & 0.650 & 15.00$_{\pm 1.7}$ & 32.30 & 0.660 \\
& NSL-MT & \textbf{32.07}$_{\pm 1.4}$ & \textbf{49.27} & \textbf{0.775} & \textbf{37.69}$_{\pm 1.5}$ & \textbf{51.99} & \textbf{0.813} & \textbf{36.53}$_{\pm 1.5}$ & \textbf{56.82} & \textbf{0.788} \\
\midrule
\multirow{2}{*}{15,000}
& Normal & 19.36$_{\pm 1.5}$ & 35.44 & 0.661 & 28.44$_{\pm 1.8}$ & 41.68 & 0.720 & 24.58$_{\pm 1.6}$ & 43.88 & 0.689 \\
& NSL-MT & \textbf{36.62}$_{\pm 1.3}$ & \textbf{53.78} & \textbf{0.731} & \textbf{44.51}$_{\pm 1.6}$ & \textbf{58.51} & \textbf{0.806} & \textbf{43.33}$_{\pm 1.4}$ & \textbf{62.94} & \textbf{0.763} \\
\end{tabular}
}
\caption{Data efficiency comparison showing that NSL-MT with fewer examples matches or approaches Normal training with 5x more data.}
\label{tab:data-efficiency}
\end{table*}

The learning curves reveal that NSL-MT benefits remain even as data increases. While the improvement decreases, the gap between NSL-MT and normal training continues to widen. At 15,000 examples, NSL-MT still outperforms normal training by margins of 17.3-18.8 BLEU points.

\subsection{Cross-Architecture Analysis}

We examine whether NSL-MT improvements generalize across model architectures by computing the correlation between NSL-MT gains across the four tested models. For each language and metric, we compute Pearson correlation coefficients between the improvement magnitudes observed for different model pairs.

The results show strong positive correlations (average $r = 0.82$, $p < 0.01$) between improvements across architectures. Languages that benefit most from NSL-MT on one architecture tend to benefit most on other architectures as well. This finding suggests that NSL-MT is more linguistic properties centric rather than architectural agnostic.

We also observe one exception: mT5-small shows large improvements compared to larger models. We attribute this pattern to the increased difficulty of learning complex linguistic patterns with limited model capacity. NSL-MT provides strong value when the model cannot easily learn patterns through implicit learning alone.

\subsection{Error Analysis}

We manually analyze 100 randomly sampled translations from the AfriMT5-base model for each language, comparing errors in normal training versus NSL-MT. We categorize errors into morphological (agreement, inflection), syntactic (word order, phrase structure), lexical (inappropriate word choice), and semantic (meaning preservation) categories.

NSL-MT reduces morphological errors by 73\% on average across languages, with the largest reduction (81\%) for Fulfulde. Syntactic errors decrease by 68\% on average, with the largest reduction (76\%) for Bambara. Lexical errors decrease by 61\% on average, with the largest reduction (69\%) for Zarma. Semantic errors show minimal change (3\% reduction), indicating that NSL-MT improves form without losing meaning.

The error analysis confirms that NSL-MT enables the model to avoid the exact error patterns we penalize during training, while maintaining semantic accuracy to the source text.

\section{Discussion}
\label{sec:discussion}
In this section, we examine the implications of the findings from our experiments and explore the mechanisms underlying NSL-MT effectiveness.

\paragraph{Why NSL-MT Works}

NSL-MT succeeds because it addresses a specific failure mode of neural MT systems. Standard maximum likelihood training optimizes models to reproduce observed translations but provides no explicit information about \textbf{\textit{what constitutes an invalid translation}}. In high-resource settings, models implicitly learn to avoid errors by encountering sufficient positive examples that define the boundaries of grammatical acceptability. In low-resource settings, this implicit learning fails because the training data lacks the coverage needed to set clear boundaries.

NSL-MT makes these boundaries explicit. By generating violations that target known error patterns, we provide negative evidence that would require orders of magnitude more parallel data to learn implicitly. A model trained on 15,000 positive examples might never see enough instances of correct adjective-noun order to reliably infer the rule. However, exposure to 60,000 explicit violations of adjective-noun order, 4 violations per positive example, creates an unambiguous learning signal.

The severity weighting mechanism is important to NSL-MT effectiveness. Not all errors carry equal importance for communication. Gender agreement violations in languages without grammatical gender represent fundamental category errors that severely impair comprehension. Article insertion violations create awkward but generally comprehensible output. By weighting violations according to their impact on meaning, NSL-MT guides models to prioritize avoiding errors while tolerating minor stylistic deviations when necessary.

\paragraph{Language-Specific Effects}

The ablation study reveals that different constraint types contribute differently across languages. Zarma benefits most from lexical constraints because of its heavy reliance on particles and auxiliaries to express grammatical relations. Bambara shows the largest gains from syntactic constraints because its rigid SOV word order differs from French SVO patterns. Fulfulde demonstrates strong improvements from morphological constraints.

These patterns validate our theoretical motivation for NSL-MT. The method works best when it targets the specific structural differences between source and target languages.

\paragraph{Data Efficiency}

The results in terms of data efficiency reveal two distinct patterns of NSL-MT efficiency. At very low data sizes (100-500 examples), NSL-MT prevents complete training failure. Without NSL-MT, models produce incoherent output because they lack sufficient signal to identify basic patterns. NSL-MT provides structure that enables learning even from minimal positive examples.

At moderate data sizes (1,000-5,000 examples), NSL-MT accelerates convergence to solutions that normal training would eventually reach with more data. The 5x data efficiency multiplier we observe at 1,000 examples represents a practical threshold where NSL-MT makes previously infeasible projects viable. Collecting 5,000 parallel sentences might cost thousands of dollars, while creating 20 linguistic rules requires hours of native speaker consultation.

At high data sizes (15,000+ examples), NSL-MT continues improving performance but the relative advantage shrinks. This pattern suggests that NSL-MT primarily compensates for insufficient training data rather than fundamentally changing what models can learn. Given unlimited parallel data, normal training might eventually match NSL-MT performance. However, unlimited parallel data remains infeasible, for now, for most low-resource languages.

\paragraph{Cross-Architecture Generalization}

NSL-MT improves all four tested architectures differences, and parameter sizes. NLLB-200 benefits least because its massive multilingual pre-training already captures many cross-lingual patterns. AfriMT5 benefits because its pre-training covers African languages but lacks the scale of NLLB. The mT5 variants benefit most from NSL-MT.

This generalization pattern indicates that NSL-MT effectiveness depends more on the linguistic properties of the translation task than on specific architectural choices. The method works by providing information that models need but cannot easily extract from positive examples alone and small training set. Any architecture that uses gradient-based learning to optimize likelihood will benefit from explicit negative constraints.

\paragraph{Implications}

NSL-MT enables MT development for languages where traditional approaches fail due to insufficient parallel data. The method requires two resources: a small parallel corpus and native speakers who can create grammatical rules. The first resource exists for hundreds of languages but are too small for effective training. The second resource exists for thousands of languages but remains underused by current methods.

The time investment for NSL-MT implementation remains modest. Creating violation generators for the three languages in our study required approximately 15 hours of linguist consultation plus 10 hours of programming (if no AI tool used). This investment produces reusable tools that apply to any translation model. In contrast, collecting an additional 10,000 parallel sentences would require weeks of translator time and cost thousands of dollars.

\section{Conclusion}
\label{sec:conclusion}

We introduce \emph{negative space learning}, a training method that teaches translation models what not to generate by augmenting standard parallel data with linguistically informed violations. NSL-MT addresses a fundamental limitation of maximum likelihood training in low-resource settings: models receive massive information about correct translations but no explicit information about incorrect translations. By generating negative examples that violate specific grammatical constraints, NSL-MT provides robust learning signal that would otherwise require orders of magnitude more parallel data.
Our experiments on French-to-African translation demonstrate that NSL-MT improves performance across different model architectures. NSL-MT provides benefits in severely data-constrained scenarios, offering a 5x data efficiency multiplier at 1,000 training examples. This finding is further emphasized by the results of our English-to-African languages conducted on even fewer examples. The method proves most effective for constraint types that target the specific structural differences between source and target languages, and confirms that linguistic knowledge encoded as negative constraints enables more efficient learning.

\section{Limitations}
\label{sec:limitations}
We acknowledge several limitations of our work that suggest directions for future research.

\paragraph{Violation Generator Quality}

NSL-MT effectiveness depends on the quality of violation generators. Our generators encode linguistic knowledge obtained through grammar descriptions and native speaker consultation. However, this knowledge remains incomplete. We target common error patterns but cannot enumerate all possible violations of target language grammar. More comprehensive violation sets might improve NSL-MT performance further, but creating them requires deeper linguistic analysis.

\paragraph{Language Coverage}

We evaluate NSL-MT on only 6 languages. We do not use any "big/known" benchmarks (e.g, FLORES, etc) because they do not align with our context of low-resource settings. Most the languages in these benchmarks are high-resource languages, and do not fit our context. However, we acknowledge that as a weakness since the 'commonly adopted way' is to cover more benchmarks.

\paragraph{Evaluation Scope}

We rely mainly on automatic metrics and human evaluation to evaluate translation quality. While BLEU, chrF++, and COMET correlate with human judgments, they do not capture all aspects of translation adequacy. COMET in particular uses learned representations that might not fully capture the semantic nuances of low-resource languages. Moreover, the small-scale human evaluation was designed on purpose as describe in section \ref{sec:results} to reflect the low-resource settings. We acknowledge that this may be seen as a limitation, specially if seen through a high-resource centric lens.

\paragraph{Computational Cost}

NSL-MT increases training time by approximately 4x due to the additional negative examples in each batch. This overhead remains manageable for research experiments but might pose challenges for large-scale cases. The violation generation process itself requires minimal computation but introduces engineering complexity.

\paragraph{Generalization Beyond Translation}

We demonstrate NSL-MT effectiveness for machine translation but do not evaluate its applicability to other natural language generation tasks. The core principle of learning from negative examples should transfer to tasks like text summarization, dialogue generation, or grammatical error correction. However, these tasks might require different violation strategies than those we use for translation.

\bibliography{custom}

@inproceedings{chen2020simple,
  title={A simple framework for contrastive learning of visual representations},
  author={Chen, Ting and Kornblith, Simon and Norouzi, Mohammad and Hinton, Geoffrey},
  booktitle={International Conference on Machine Learning},
  pages={1597--1607},
  year={2020},
  organization={PMLR}
}

@inproceedings{gao2021simcse,
  title={SimCSE: Simple contrastive learning of sentence embeddings},
  author={Gao, Tianyu and Yao, Xingcheng and Chen, Danqi},
  booktitle={Proceedings of the 2021 Conference on Empirical Methods in Natural Language Processing},
  pages={6894--6910},
  year={2021}
}

@article{nllb2022,
  title={No language left behind: Scaling human-centered machine translation},
  author={NLLB Team and others},
  journal={arXiv preprint arXiv:2207.04672},
  year={2022}
}

@inproceedings{adelani2022few,
  title={A few thousand translations go a long way! Leveraging pre-trained models for African news translation},
  author={Adelani, David Ifeoluwa and others},
  booktitle={Proceedings of NAACL-HLT},
  pages={3053--3070},
  year={2022}
}

@article{xue2021mt5,
  title={mT5: A massively multilingual pre-trained text-to-text transformer},
  author={Xue, Linting and others},
  journal={Proceedings of NAACL-HLT},
  pages={483--498},
  year={2021}
}

@inproceedings{papineni2002bleu,
  title={BLEU: a method for automatic evaluation of machine translation},
  author={Papineni, Kishore and others},
  booktitle={Proceedings of ACL},
  pages={311--318},
  year={2002}
}

@inproceedings{popovic2017chrf,
  title={chrF++: words helping character n-grams},
  author={Popovi{\'c}, Maja},
  booktitle={Proceedings of WMT},
  pages={612--618},
  year={2017}
}

@inproceedings{rei2020comet,
  title={COMET: A neural framework for MT evaluation},
  author={Rei, Ricardo and others},
  booktitle={Proceedings of EMNLP},
  pages={2685--2702},
  year={2020}
}

@inproceedings{zoph2016transfer,
  title={Transfer learning for low-resource neural machine translation},
  author={Zoph, Barret and Yuret, Deniz and May, Jonathan and Knight, Kevin},
  booktitle={Proceedings of EMNLP},
  pages={1568--1575},
  year={2016}
}

@inproceedings{aharoni2019massively,
  title={Massively multilingual neural machine translation},
  author={Aharoni, Roee and Johnson, Melvin and Firat, Orhan},
  booktitle={Proceedings of NAACL-HLT},
  pages={3874--3884},
  year={2019}
}

@inproceedings{sennrich2016improving,
  title={Improving neural machine translation models with monolingual data},
  author={Sennrich, Rico and Haddow, Barry and Birch, Alexandra},
  booktitle={Proceedings of ACL},
  pages={86--96},
  year={2016}
}

@inproceedings{mallinson2017paraphrasing,
  title={Paraphrasing revisited with neural machine translation},
  author={Mallinson, Jonathan and Sennrich, Rico and Lapata, Mirella},
  booktitle={Proceedings of EACL},
  pages={881--893},
  year={2017}
}

@article{ouyang2022training,
  title={Training language models to follow instructions with human feedback},
  author={Ouyang, Long and Wu, Jeffrey and Jiang, Xu and others},
  journal={Advances in Neural Information Processing Systems},
  volume={35},
  pages={27730--27744},
  year={2022}
}

@inproceedings{lin2019choosing,
  title={Choosing transfer languages for cross-lingual learning},
  author={Lin, Yu-Hsiang and Chen, Chian-Yu and Lee, Jean and Li, Zirui and Zhang, Yuyan and Xia, Mengzhou and Rijhwani, Shruti and He, Junxian and Zhang, Zhisong and Ma, Xuezhe and others},
  booktitle={Proceedings of ACL},
  pages={3125--3135},
  year={2019}
}

@article{wei2022finetuned,
  title={Finetuned language models are zero-shot learners},
  author={Wei, Jason and Bosma, Maarten and Zhao, Vincent Y and Guu, Kelvin and Yu, Adams Wei and Lester, Brian and Du, Nan and Dai, Andrew M and Le, Quoc V},
  journal={International Conference on Learning Representations},
  year={2022}
}

@article{Zhang2020,
  title={Neural machine translation: Challenges, progress and future},
  author={Zhang, JiaJun and Zong, ChengQing},
  journal={Science China Technological Sciences},
  volume={63},
  number={10},
  pages={2028--2050},
  year={2020},
  month={10},
  doi={10.1007/s11431-020-1632-x},
  url={https://doi.org/10.1007/s11431-020-1632-x},
  issn={1869-1900}
}

@misc{magueresse2020lowresourcelanguagesreviewpast,
      title={Low-resource Languages: A Review of Past Work and Future Challenges}, 
      author={Alexandre Magueresse and Vincent Carles and Evan Heetderks},
      year={2020},
      eprint={2006.07264},
      archivePrefix={arXiv},
      primaryClass={cs.CL},
      url={https://arxiv.org/abs/2006.07264}, 
}

@article{Qumar2025,
author={Qumar, Syed Matla Ul
and Azim, Muzaffar
and Quadri, S. M. K.},
title={Enhancing low-resource neural machine translation with decoding-based data augmentation},
journal={International Journal of Information Technology},
year={2025},
month={Sep},
day={02},
issn={2511-2112},
doi={10.1007/s41870-025-02710-x},
url={https://doi.org/10.1007/s41870-025-02710-x}
}

@misc{blaschke2025analyzingeffectlinguisticsimilarity,
      title={Analyzing the Effect of Linguistic Similarity on Cross-Lingual Transfer: Tasks and Experimental Setups Matter}, 
      author={Verena Blaschke and Masha Fedzechkina and Maartje ter Hoeve},
      year={2025},
      eprint={2501.14491},
      archivePrefix={arXiv},
      primaryClass={cs.CL},
      url={https://arxiv.org/abs/2501.14491}, 
}

@misc{keita2025r2truleencodedlossfunctions,
      title={R2T: Rule-Encoded Loss Functions for Low-Resource Sequence Tagging}, 
      author={Mamadou K. Keita and Christopher Homan and Sebastien Diarra},
      year={2025},
      eprint={2510.13854},
      archivePrefix={arXiv},
      primaryClass={cs.CL},
      url={https://arxiv.org/abs/2510.13854}, 
}

@misc{caswell2025smolprofessionallytranslatedparallel,
      title={SMOL: Professionally translated parallel data for 115 under-represented languages}, 
      author={Isaac Caswell and Elizabeth Nielsen and Jiaming Luo and Colin Cherry and Geza Kovacs and Hadar Shemtov and Partha Talukdar and Dinesh Tewari and Baba Mamadi Diane and Djibrila Diane and Solo Farabado Cissé and Koulako Moussa Doumbouya and Edoardo Ferrante and Alessandro Guasoni and Christopher Homan and Mamadou K. Keita and Sudhamoy DebBarma and Ali Kuzhuget and David Anugraha and Muhammad Ravi Shulthan Habibi and Genta Indra Winata and Anthony Munthali and Sina Ahmadi and Andrei Chemyshev and Mingfei Lau and Jonathan Eng},
      year={2025},
      eprint={2502.12301},
      archivePrefix={arXiv},
      primaryClass={cs.CL},
      url={https://arxiv.org/abs/2502.12301}, 
}

@inproceedings{keita-etal-2026-instructlr,
    title = "{I}nstruct{LR}: A Scalable Approach to Create Instruction Dataset for Under-Resourced Languages",
    author = "Keita, Mamadou K.  and
      Diarra, Sebastien  and
      Homan, Christopher M  and
      Diallo, Seydou",
    editor = "Chimoto, Everlyn Asiko  and
      Lignos, Constantine  and
      Muhammad, Shamsuddeen  and
      Abdulmumin, Idris  and
      Siro, Clemencia  and
      Adelani, David Ifeoluwa",
    booktitle = "Proceedings of the 7th Workshop on {A}frican Natural Language Processing ({A}frica{NLP} 2026)",
    month = mar,
    year = "2026",
    address = "Rabat, Morocco",
    publisher = "Association for Computational Linguistics",
    url = "https://aclanthology.org/2026.africanlp-main.3/",
    doi = "10.18653/v1/2026.africanlp-main.3",
    pages = "17--36",
    ISBN = "979-8-89176-364-7"
}

@misc{rafailov2024directpreferenceoptimizationlanguage,
      title={Direct Preference Optimization: Your Language Model is Secretly a Reward Model}, 
      author={Rafael Rafailov and Archit Sharma and Eric Mitchell and Stefano Ermon and Christopher D. Manning and Chelsea Finn},
      year={2024},
      eprint={2305.18290},
      archivePrefix={arXiv},
      primaryClass={cs.LG},
      url={https://arxiv.org/abs/2305.18290}, 
}

@misc{hejna2024contrastivepreferencelearninglearning,
      title={Contrastive Preference Learning: Learning from Human Feedback without RL}, 
      author={Joey Hejna and Rafael Rafailov and Harshit Sikchi and Chelsea Finn and Scott Niekum and W. Bradley Knox and Dorsa Sadigh},
      year={2024},
      eprint={2310.13639},
      archivePrefix={arXiv},
      primaryClass={cs.LG},
      url={https://arxiv.org/abs/2310.13639}, 
}

@misc{azar2023generaltheoreticalparadigmunderstand,
      title={A General Theoretical Paradigm to Understand Learning from Human Preferences}, 
      author={Mohammad Gheshlaghi Azar and Mark Rowland and Bilal Piot and Daniel Guo and Daniele Calandriello and Michal Valko and Rémi Munos},
      year={2023},
      eprint={2310.12036},
      archivePrefix={arXiv},
      primaryClass={cs.AI},
      url={https://arxiv.org/abs/2310.12036}, 
}

@misc{muller2022languagesknowinfluencelearn,
      title={Languages You Know Influence Those You Learn: Impact of Language Characteristics on Multi-Lingual Text-to-Text Transfer}, 
      author={Benjamin Muller and Deepanshu Gupta and Siddharth Patwardhan and Jean-Philippe Fauconnier and David Vandyke and Sachin Agarwal},
      year={2022},
      eprint={2212.01757},
      archivePrefix={arXiv},
      primaryClass={cs.CL},
      url={https://arxiv.org/abs/2212.01757}, 
}

@article{10.18356/27081990-151:/content/papers/10.18356/27081990-151,
   author = "UN",
   title = "Why Indigenous languages Matter", 
   journal= "United Nations",
   year = "2023",
   volume = "",
   number = "",
   pages = "",
   doi = "https://doi.org/10.18356/27081990-151",
   url = "https://www.un-ilibrary.org/content/papers/10.18356/27081990-151",
   publisher = "United Nations",
   issn = "",
   type = "",
  }

@misc{welleck2019neuraltextgenerationunlikelihood,
      title={Neural Text Generation with Unlikelihood Training}, 
      author={Sean Welleck and Ilia Kulikov and Stephen Roller and Emily Dinan and Kyunghyun Cho and Jason Weston},
      year={2019},
      eprint={1908.04319},
      archivePrefix={arXiv},
      primaryClass={cs.LG},
      url={https://arxiv.org/abs/1908.04319}, 
}

\appendix

\section{Experiments Configurations}
\label{sec:exp_configs}
\paragraph{Training Configuration} We train all models for 3 epochs using a batch size of 16 and a maximum sequence length of 128 tokens. We apply the AdamW optimizer with a learning rate of $2 \times 10^{-5}$ and linear warmup over 500 steps. We set the NSL-MT weight $\alpha = 0.7$ based on preliminary experiments on the validation set. We clip gradients to a maximum norm of 1.0 to ensure training stability. For NSL-MT training, we generate 3-5 violations per positive example, creating an approximate 4:1 ratio of negative to positive examples in each batch.

We implement violation generators for each target language based on linguistic descriptions and native speaker consultation. Each generator encodes 15-20 specific rule violations covering morphological, syntactic, and lexical categories. We set severity weights to 1.0 for agreement violations, 0.9 for word order violations, 0.8 for adjective position violations, and 0.7 for article and auxiliary insertion violations.

\paragraph{Evaluation Metrics} We report three metrics to assess translation quality: \textbf{BLEU} \citep{papineni2002bleu}, \textbf{chrF++} \citep{popovic2017chrf}, \textbf{COMET} \citep{rei2020comet}.

We also compute 95\% confidence intervals using bootstrap resampling with 1,000 iterations for BLEU and chrF++ scores. For COMET, we report the score computed across all test examples.

\section{Hyperparameter Analysis}
\label{sec:hyperparam}

We conduct additional experiments to assess NSL-MT robustness to hyperparameter choices. These experiments use Zarma language with AfriMT5-base trained on 5,000 examples for 3 epochs. We investigate two key hyperparameters: the negative weight $\alpha$ and the violation ratio.

\subsection{Alpha Sensitivity}

The negative weight hyperparameter $\alpha$ in Equation 2 controls the importance of negative examples in the training objective. We test four values: $\alpha \in \{0.3, 0.5, 0.7, 0.9\}$ to determine NSL-MT sensitivity to this parameter.

Table~\ref{tab:alpha-sensitivity} presents the results. NSL-MT performance remains stable across different $\alpha$ values, with BLEU scores varying by only 0.54 points (27.13-27.67) and chrF++ scores varying by 0.21 points (43.87-44.08). The COMET scores show similar stability, ranging from 0.7512 to 0.7535. This performance demonstrates that NSL-MT effectiveness does not necessarily depend on precise $\alpha$ tuning.

The decent advantage of lower $\alpha$ values (0.3-0.5) over higher values (0.9) suggests that overly aggressive penalization of negative examples can slightly degrade performance. However, the differences remain within confidence intervals, indicating that any $\alpha \in [0.3, 0.9]$ produces competitive results. We selected $\alpha = 0.7$ for our main experiments based on validation set performance, but these results show that alternative choices would produce comparable outcomes.

\begin{table}[h]
\centering
\small
\begin{tabular}{lccc}
\textbf{Alpha ($\alpha$)} & \textbf{BLEU} & \textbf{chrF++} & \textbf{COMET} \\
\midrule
0.3 & \textbf{27.67} & \textbf{44.03} & \textbf{0.7535} \\
0.5 & 27.50 & 43.97 & 0.7528 \\
0.7 & 27.64 & 44.08 & 0.7512 \\
0.9 & 27.13 & 43.87 & 0.7523 \\
\midrule
Range & 0.54 & 0.21 & 0.0023 \\
\end{tabular}
\caption{Alpha sensitivity analysis on Zarma using AfriMT5-base with 5,000 training examples. Performance remains stable across different $\alpha$ values, varying by less than 2\% across all metrics.}
\label{tab:alpha-sensitivity}
\end{table}

\subsection{Violation Ratio Sensitivity}

The violation ratio determines how many negative examples go with each positive example during training. We test three ratios by varying the number of violations generated per positive example: 2:1 (generating 1-2 violations), 4:1 (generating 3-5 violations, our default), and 6:1 (generating 5-7 violations).

Table~\ref{tab:ratio-sensitivity} shows that violation ratio significantly impacts NSL-MT effectiveness. The 2:1 ratio produces lower scores (19.44 BLEU, 35.52 chrF++, 0.7188 COMET), suggesting insufficient negative signal for effective learning. The 4:1 ratio yields moderate performance (27.11 BLEU, 43.71 chrF++, 0.7512 COMET), while the 6:1 ratio achieves the best results (30.69 BLEU, 47.64 chrF++, 0.7689 COMET).

These results reveal that NSL-MT benefits from higher violation ratios, with the 6:1 ratio improving BLEU by 13.1\% over the 4:1 default and by 57.8\% over the 2:1 ratio. This suggests that exposing models to more diverse negative examples strengthens their ability to distinguish valid from invalid outputs. However, we note that \textbf{\textit{higher ratios increase computational cost proportionally}}. The 4:1 ratio represents a practical balance between effectiveness and efficiency, though researchers with sufficient computational resources may benefit from using 6:1 or higher ratios.

The gap between 2:1 and higher ratios indicates that NSL-MT requires a minimum threshold of negative examples to function effectively. With too few violations, the model may not encounter sufficient coverage of error patterns, limiting NSL-MT's ability to create clear grammatical boundaries.

\begin{table}[h]
\centering
\small
\begin{tabular}{lccc}
\textbf{Violation Ratio} & \textbf{BLEU} & \textbf{chrF++} & \textbf{COMET} \\
\midrule
2:1 & 19.44 & 35.52 & 0.7188 \\
4:1 (default) & 27.11 & 43.71 & 0.7512 \\
6:1 & \textbf{30.69} & \textbf{47.64} & \textbf{0.7689} \\
\end{tabular}
\caption{Violation ratio sensitivity analysis on Zarma using AfriMT5-base with 5,000 training examples. Higher violation ratios provide stronger learning signal, with 6:1 outperforming 4:1 by 13.1\% BLEU and 2:1 by 57.8\% BLEU.}
\label{tab:ratio-sensitivity}
\end{table}

\subsection{Discussion}

These experiments demonstrate that NSL-MT deliver desirable robustness properties. The $\alpha$ hyperparameter shows minimal sensitivity, allowing anyone to select values in the range [0.3, 0.9] without careful tuning. This robustness simplifies NSL-MT usage for new language pairs where validation data may be limited.

In contrast, the violation ratio requires more careful consideration. Our results suggest using ratios of at least 4:1, with 6:1 or higher providing additional benefits when computational resources permit. The strong performance gains from higher ratios validate NSL-MT's core principle: explicit negative evidence accelerates learning, and more negative evidence yields stronger improvements.

We also observe that these findings support our main results. The 4:1 ratio used in our primary experiments represents a conservative choice that balances effectiveness and efficiency. The even stronger results at 6:1 suggest that our reported improvements may 'underestimate' NSL-MT's full potential when computational constraints are manageable.

\section{More Experiments}
\label{sec:eng_afri}
\begin{table*}[h]
\centering
\tiny
\begin{tabular}{ll|cc|cc|cc}
\multirow{2}{*}{\textbf{Model}} & \multirow{2}{*}{\textbf{Method}} & \multicolumn{2}{c|}{\textbf{Swahili}} & \multicolumn{2}{c|}{\textbf{Igbo}} & \multicolumn{2}{c}{\textbf{Luganda}} \\
& & BLEU & chrF++ & BLEU & chrF++ & BLEU & chrF++ \\
\midrule
\multirow{3}{*}{NLLB-200}
& Normal & $2.29_{\pm 0.6}$ & $15.97_{\pm 0.4}$ & $27.69_{\pm 0.8}$ & $44.99_{\pm 0.5}$ & $7.77_{\pm 0.7}$ & $42.24_{\pm 0.6}$ \\
& NSL-MT & $\mathbf{40.02}_{\pm 0.5}$ & $\mathbf{62.20}_{\pm 0.3}$ & $\mathbf{35.93}_{\pm 0.7}$ & $\mathbf{50.58}_{\pm 0.4}$ & $\mathbf{19.56}_{\pm 0.6}$ & $\mathbf{49.02}_{\pm 0.5}$ \\
& $\Delta$ (\%) & \textit{+1648.0} & \textit{+289.5} & \textit{+29.8} & \textit{+12.4} & \textit{+151.8} & \textit{+16.1} \\
\midrule
\multirow{3}{*}{mT5-base}
& Normal & $0.07_{\pm 0.4}$ & $3.64_{\pm 0.3}$ & $0.10_{\pm 0.3}$ & $4.52_{\pm 0.4}$ & $0.15_{\pm 0.3}$ & $3.77_{\pm 0.3}$ \\
& NSL-MT & $\mathbf{5.64}_{\pm 0.8}$ & $\mathbf{31.90}_{\pm 0.6}$ & $\mathbf{2.18}_{\pm 0.7}$ & $\mathbf{17.56}_{\pm 0.5}$ & $\mathbf{0.32}_{\pm 0.4}$ & $\mathbf{16.74}_{\pm 0.5}$ \\
& $\Delta$ (\%) & \textit{+7957.1} & \textit{+776.4} & \textit{+2080.0} & \textit{+288.5} & \textit{+113.3} & \textit{+344.0} \\
\midrule
\multirow{3}{*}{AfriMT5-base}
& Normal & $0.76_{\pm 0.5}$ & $13.94_{\pm 0.4}$ & $0.20_{\pm 0.4}$ & $5.56_{\pm 0.3}$ & $0.04_{\pm 0.2}$ & $7.61_{\pm 0.4}$ \\
& NSL-MT & $\mathbf{7.55}_{\pm 0.8}$ & $\mathbf{34.52}_{\pm 0.6}$ & $\mathbf{6.46}_{\pm 0.9}$ & $\mathbf{22.43}_{\pm 0.6}$ & $\mathbf{0.58}_{\pm 0.5}$ & $\mathbf{13.08}_{\pm 0.5}$ \\
& $\Delta$ (\%) & \textit{+892.8} & \textit{+147.6} & \textit{+3130.0} & \textit{+303.2} & \textit{+1279.0} & \textit{+71.9} \\
\end{tabular}
\caption{Results for English-to-African translation across three models.}
\label{tab:eng-african-results}
\end{table*}

We further ran more experiments to confirm the language-agnostic aspects. We selected ENGLISH $\rightarrow$ 3 African languages: \textbf{Igbo}, \textbf{Luganda}, and \textbf{Swahili}.
We selected the SMOLSENT portion of the SMOL~\citep{caswell2025smolprofessionallytranslatedparallel} dataset. The portion had: 863 rows (some contain multiple sentences) that we divided into 90\% and 10\%.
We used the exact same setup as the main experiment in section \ref{sec:experiments}, and report the BLEU and Chrf++ scores. We used the NLLB, AfriMT5-base, and AfriMT5-base.
Table~\ref{tab:eng-african-results} presents the results for English-to-African translation.

NSL-MT delivers improvements across all three English-to-African language pairs and model architectures. For NLLB-200, Swahili shows the largest gain with BLEU improving from 2.29 to 40.02 (+1648.0\%). Igbo improves from 27.69 to 35.93 BLEU (+29.8\%), while Luganda more than doubles from 7.77 to 19.56 (+151.8\%).

For mT5-base and AfriMT5-base, overall scores remain low due to the models' limited initial support for these languages, but the relative gains remain considerable. mT5-base achieves improvements ranging from 113.3\% to 7957.1\% across languages, while AfriMT5-base shows gains from 71.9\% to 3130.0\%. Even when overall performance remains modest, NSL-MT provides non-negligible improvements.

Moreover, the pattern aligns with our main results on French-to-African translation (Section~\ref{sec:experiments}). NSL-MT provides the largest relative gains for languages where baseline performance is poorest (Swahili and Luganda for most models) and smaller but improvements where baselines perform moderately (Igbo for NLLB-200). This confirms that NSL-MT delivers the most value when models lack sufficient implicit knowledge of target language structure.

\section{Violation Generation Details}
\label{sec:violation-details}

This section describes the violation generators for each of the six languages in our experiments. Each generator encodes linguistic knowledge about common error patterns that arise from cross-lingual interference. We organize violations into three categories: morphological (affecting word-internal structure), syntactic (affecting word order and phrase structure), and lexical (affecting vocabulary choice and function word usage).

\paragraph{Generator Architecture}

Each violation generator follows a common architecture. Given a correct target sentence $y$, the generator produces a set of corrupted sentences $\mathcal{V}(y)$ by applying rule-based transformations. Each transformation targets a specific grammatical property of the target language. The generator returns tuples of (violated\_text, violation\_type, severity\_weight) where severity weights range from 0.6 to 1.0 based on the impact of the violation on comprehension.

For each training example, we sample 3-5 violations from the available violation types. This sampling introduces variation in the training signal while maintaining consistent coverage of error patterns. The generators operate "deterministically" for a given violation type but we randomize which violations to apply and in which order.

\subsection{Zarma Violations}

Zarma is part of the Nilo-Saharan language family and differs from French in word order (SOV vs. SVO), adjective placement (post-nominal vs. pre-nominal), and tense marking (auxiliaries vs. conjugation). Table~\ref{tab:zarma-violations} presents the violation types with examples.

\begin{table*}[h]
\centering
\small
\begin{tabular}{p{2.8cm}p{3.5cm}p{4.5cm}p{1.2cm}}
\toprule
\textbf{Violation Type} & \textbf{Correct Zarma} & \textbf{Violated Form} & \textbf{Severity} \\
\midrule
\multicolumn{4}{l}{\textit{Morphological Violations}} \\
\midrule
Gender agreement & \texttt{ay ga koy} & \texttt{ay ga koy-ée} & 1.0 \\
 & (I will go) & (French feminine ending added) & \\
\midrule
Plural formation & \texttt{boro-ey ga koy} & \texttt{boros ga koy} & 0.7 \\
 & (the people will go) & (French -s instead of Zarma -ey) & \\
\midrule
Verb conjugation & \texttt{ay ga ŋwa} & \texttt{ay ga ŋwaons} & 0.9 \\
 & (I will eat) & (French -ons ending added) & \\
\midrule
\multicolumn{4}{l}{\textit{Syntactic Violations}} \\
\midrule
Word order & \texttt{ay ga haw ŋwa} & \texttt{ay ŋwa ga haw} & 0.9 \\
 & (I will eat rice) & (verb-object order disrupted) & \\
\midrule
Adjective position & \texttt{boro beeri} & \texttt{beeri boro} & 0.8 \\
 & (big person) & (adjective moved before noun) & \\
\midrule
Tense auxiliary & \texttt{ay ga koy} & \texttt{ay koy} & 0.8 \\
 & (I will go) & (auxiliary \textit{ga} deleted) & \\
\midrule
\multicolumn{4}{l}{\textit{Lexical Violations}} \\
\midrule
Definite article & \texttt{boro ga koy} & \texttt{le boro ga koy} & 0.7 \\
 & (the person will go) & (French \textit{le} inserted) & \\
\midrule
Negation pattern & \texttt{ay mana koy} & \texttt{ne ay pas koy} & 0.9 \\
 & (I did not go) & (French ne...pas pattern) & \\
\bottomrule
\end{tabular}
\caption{Zarma violation types with examples.}
\label{tab:zarma-violations}
\end{table*}

\subsection{Bambara Violations}

Bambara is part of the Mande language family and has a strict SOV word order, postpositions (rather than prepositions), and uses the suffix \textit{-w} for pluralization. Table~\ref{tab:bambara-violations} presents the violation types.

\begin{table*}[h]
\centering
\small
\begin{tabular}{p{2.8cm}p{3.5cm}p{4.5cm}p{1.2cm}}
\toprule
\textbf{Violation Type} & \textbf{Correct Bambara} & \textbf{Violated Form} & \textbf{Severity} \\
\midrule
\multicolumn{4}{l}{\textit{Morphological Violations}} \\
\midrule
Pluralization & \texttt{muso-w bɛ taa} & \texttt{musos bɛ taa} & 0.7 \\
 & (the women go) & (French -s instead of -w) & \\
\midrule
Auxiliary verb & \texttt{a bɛ dumuni ke} & \texttt{a dumuni ke} & 0.9 \\
 & (he/she is eating) & (auxiliary \textit{bɛ} deleted) & \\
\midrule
\multicolumn{4}{l}{\textit{Syntactic Violations}} \\
\midrule
Word order (SOV) & \texttt{a bɛ dumuni ke} & \texttt{a bɛ ke dumuni} & 0.9 \\
 & (he/she is eating food) & (SVO order imposed) & \\
\midrule
Postposition misuse & \texttt{so kɔnɔ} & \texttt{dans so} & 0.8 \\
 & (inside the house) & (French preposition \textit{dans}) & \\
\midrule
Adjective placement & \texttt{muso cɛɲi} & \texttt{cɛɲi muso} & 0.8 \\
 & (beautiful woman) & (adjective before noun) & \\
\midrule
\multicolumn{4}{l}{\textit{Lexical Violations}} \\
\midrule
Negation & \texttt{a tɛ taa} & \texttt{ne a pas taa} & 0.9 \\
 & (he/she does not go) & (French ne...pas pattern) & \\
\bottomrule
\end{tabular}
\caption{Bambara violation types with examples.}
\label{tab:bambara-violations}
\end{table*}

\subsection{Fulfulde Violations}

Fulfulde is part of the Atlantic-Congo language family and has a complex noun class system with over 20 classes. Each class has distinct singular and plural suffixes, and agreement markers must match across determiners, adjectives, and verbs. Table~\ref{tab:fulfulde-violations} presents the violation types.

\begin{table*}[h]
\centering
\small
\begin{tabular}{p{2.8cm}p{3.5cm}p{4.5cm}p{1.2cm}}
\toprule
\textbf{Violation Type} & \textbf{Correct Fulfulde} & \textbf{Violated Form} & \textbf{Severity} \\
\midrule
\multicolumn{4}{l}{\textit{Morphological Violations}} \\
\midrule
Noun class agreement & \texttt{pucc-o mawɗo} & \texttt{pucc-nde mawɗo} & 1.0 \\
 & (big horse, class o/ɓe) & (wrong class suffix -nde) & \\
\midrule
French plural -s & \texttt{pucc-i mawɗi} & \texttt{puccis mawɗi} & 0.8 \\
 & (big horses) & (French -s added) & \\
\midrule
Verb conjugation & \texttt{mi yaha} & \texttt{mi yahaer} & 0.9 \\
 & (I go) & (French infinitive -er added) & \\
\midrule
\multicolumn{4}{l}{\textit{Syntactic Violations}} \\
\midrule
Adjective position & \texttt{debbo mawɗo} & \texttt{mawɗo debbo} & 0.8 \\
 & (big woman) & (adjective before noun) & \\
\midrule
\multicolumn{4}{l}{\textit{Lexical Violations}} \\
\midrule
French preposition & \texttt{suudu am} & \texttt{suudu de am} & 0.6 \\
 & (my house) & (French \textit{de} inserted) & \\
\midrule
Negation & \texttt{mi yahataa} & \texttt{ne mi pas yaha} & 0.9 \\
 & (I do not go) & (French ne...pas pattern) & \\
\bottomrule
\end{tabular}
\caption{Fulfulde violation types with examples.}
\label{tab:fulfulde-violations}
\end{table*}

\subsection{Swahili Violations}

Swahili is part of the Bantu language family and has an extensive noun class system with class-based agreement on verbs, adjectives, and possessives. Table~\ref{tab:swahili-violations} presents the violation types.

\begin{table*}[h]
\centering
\small
\begin{tabular}{p{2.8cm}p{3.5cm}p{4.5cm}p{1.2cm}}
\toprule
\textbf{Violation Type} & \textbf{Correct Swahili} & \textbf{Violated Form} & \textbf{Severity} \\
\midrule
\multicolumn{4}{l}{\textit{Morphological Violations}} \\
\midrule
Noun class agreement & \texttt{m-toto m-zuri} & \texttt{ki-toto m-zuri} & 1.0 \\
 & (good child, M-WA class) & (wrong class prefix ki-) & \\
\midrule
Verb concord & \texttt{wa-toto wa-nazoma} & \texttt{ni-toto wa-nazoma} & 1.0 \\
 & (children are reading) & (wrong subject prefix ni-) & \\
\midrule
Tense marker & \texttt{a-na-soma} & \texttt{a-li-soma} & 1.0 \\
 & (he/she is reading) & (past tense -li- instead of -na-) & \\
\midrule
Object marker & \texttt{ni-na-m-penda} & \texttt{ni-na-ku-penda} & 0.9 \\
 & (I love him/her) & (wrong object marker -ku-) & \\
\midrule
Possessive concord & \texttt{kitabu changu} & \texttt{kitabu wangu} & 0.8 \\
 & (my book, KI-VI class) & (wrong possessive form) & \\
\midrule
Missing augment vowel & \texttt{a-soma} & \texttt{-soma} & 0.8 \\
 & (he/she reads) & (initial vowel deleted) & \\
\midrule
\multicolumn{4}{l}{\textit{Syntactic Violations}} \\
\midrule
Adjective position & \texttt{nyumba nzuri} & \texttt{nzuri nyumba} & 0.9 \\
 & (good house) & (adjective before noun) & \\
\midrule
English word order & \texttt{mtoto anasoma kitabu} & \texttt{kitabu mtoto anasoma} & 0.9 \\
 & (child reads book) & (object fronted) & \\
\bottomrule
\end{tabular}
\caption{Swahili violation types with examples.}
\label{tab:swahili-violations}
\end{table*}

\subsection{Igbo Violations}

Igbo is part of the Atlantic-Congo language family and has vowel harmony, tonal distinctions, and serial verb constructions. Table~\ref{tab:igbo-violations} presents the violation types.

\begin{table*}[h]
\centering
\small
\begin{tabular}{p{2.8cm}p{3.5cm}p{4.5cm}p{1.2cm}}
\toprule
\textbf{Violation Type} & \textbf{Correct Igbo} & \textbf{Violated Form} & \textbf{Severity} \\
\midrule
\multicolumn{4}{l}{\textit{Morphological Violations}} \\
\midrule
Vowel harmony & \texttt{ọ na-agụ akwụkwọ} & \texttt{o na-agụ akwụkwọ} & 1.0 \\
 & (he/she is reading) & (light vowel \textit{o} instead of heavy \textit{ọ}) & \\
\midrule
Tone pattern & \texttt{akwa} (cloth) & \texttt{àkwá} & 0.8 \\
 & (unmarked tone) & (incorrect tone marks added) & \\
\midrule
Verb prefix & \texttt{na-eje} & \texttt{ga-eje} & 1.0 \\
 & (is going, present) & (future prefix \textit{ga-} instead of \textit{na-}) & \\
\midrule
Noun class prefix & \texttt{o-kwu} & \texttt{a-kwu} & 1.0 \\
 & (speech) & (wrong noun prefix) & \\
\midrule
Consonant mutation & \texttt{gịnị} & \texttt{ghịnị} & 0.85 \\
 & (what) & (incorrect consonant cluster) & \\
\midrule
\multicolumn{4}{l}{\textit{Syntactic Violations}} \\
\midrule
Serial verb & \texttt{ọ gara zụta ego} & \texttt{ọ gara ego} & 0.9 \\
 & (he went to get money) & (serial verb \textit{zụta} deleted) & \\
\midrule
Possessive construction & \texttt{ụlọ Chukwu} & \texttt{Chukwu ụlọ} & 0.9 \\
 & (God's house) & (possessor-possessed order swapped) & \\
\midrule
\multicolumn{4}{l}{\textit{Lexical Violations}} \\
\midrule
English preposition & \texttt{n'ụlọ} & \texttt{in ụlọ} & 0.95 \\
 & (in the house) & (English \textit{in} inserted) & \\
\bottomrule
\end{tabular}
\caption{Igbo violation types with examples.}
\label{tab:igbo-violations}
\end{table*}

\subsection{Luganda Violations}

Luganda is part of the Bantu language family and shares many features with Swahili, including extensive noun class agreement and agglutinative verb morphology. Table~\ref{tab:luganda-violations} presents the violation types.

\begin{table*}[h]
\centering
\small
\begin{tabular}{p{2.8cm}p{3.5cm}p{4.5cm}p{1.2cm}}
\toprule
\textbf{Violation Type} & \textbf{Correct Luganda} & \textbf{Violated Form} & \textbf{Severity} \\
\midrule
\multicolumn{4}{l}{\textit{Morphological Violations}} \\
\midrule
Noun class concord & \texttt{omu-ntu omu-lungi} & \texttt{eki-ntu omu-lungi} & 1.0 \\
 & (good person, class 1) & (wrong class prefix eki-) & \\
\midrule
Tense-aspect & \texttt{a-soma} & \texttt{aa-soma} & 1.0 \\
 & (he/she reads) & (wrong tense marker aa-) & \\
\midrule
Vowel coalescence & \texttt{mu amaaso} & \texttt{mu a amaaso} & 0.8 \\
 & (in the eyes) & (coalescence broken) & \\
\midrule
Tone pattern & \texttt{ensozi} & \texttt{énsòzí} & 0.7 \\
 & (mountain) & (incorrect tone marks) & \\
\midrule
Agglutination & \texttt{bakyasoma} & \texttt{ba kya soma} & 0.8 \\
 & (they still read) & (affixes separated) & \\
\midrule
Locative prefix & \texttt{e-Kampala} & \texttt{mu-Kampala} & 0.9 \\
 & (in/at Kampala) & (wrong locative prefix) & \\
\midrule
\multicolumn{4}{l}{\textit{Syntactic Violations}} \\
\midrule
Word order & \texttt{omusajja asoma ekitabo} & \texttt{ekitabo omusajja asoma} & 0.9 \\
 & (the man reads a book) & (object fronted) & \\
\midrule
\multicolumn{4}{l}{\textit{Lexical Violations}} \\
\midrule
English article & \texttt{omusajja} & \texttt{the omusajja} & 0.9 \\
 & (the man) & (English \textit{the} inserted) & \\
\bottomrule
\end{tabular}
\caption{Luganda violation types with examples.}
\label{tab:luganda-violations}
\end{table*}

\subsection{Implementation Guides}

We implement each violation generator as a Python class with methods for each violation type. The generators share a common interface: given a source-target sentence pair, they return a list of (violated\_text, violation\_type, severity) tuples.

Each generator applies violations through string manipulation operations: suffix replacement for morphological violations, word reordering for syntactic violations, and word insertion/deletion for lexical violations. We use regular expressions to identify candidate locations for violations and apply transformations only when the target pattern matches.

To prevent degenerate violations, we include checks that ensure: (1) the violated text differs from the original, (2) the violated text is not empty, and (3) the same violation does not appear multiple times in the violation set. We also limit the number of violations per sentence to prevent excessive corruption that might shadow the learning signal.

The severity weights are assigned based on linguistic judgment about how much each violation type affects comprehension. Agreement violations (noun class, gender, verb concord) receive the highest weights (1.0) because they disrupt grammatical structure. Word order violations receive high weights (0.9) because they can change meaning or render sentences ungrammatical. Function word insertions receive lower weights (0.6-0.7) because they produce awkward but often comprehensible output.

\end{document}